\documentclass[letterpaper]{article} 
\usepackage{aaai21}  
\usepackage{times}  
\usepackage{helvet} 
\usepackage{courier}  
\usepackage[hyphens]{url}  
\usepackage{graphicx} 
\usepackage{natbib}  
\usepackage{caption} 
\usepackage{amsmath,amssymb} 
\usepackage{multirow} 

\usepackage{booktabs}
\usepackage{xspace}

\usepackage[ruled,linesnumbered]{algorithm2e}

\newcommand{\be}{\mathbf e}

\newcommand{\bW}{\mathbf W}
\let\oldnl\nl
\newcommand{\nonl}{\renewcommand{\nl}{\let\nl\oldnl}}

\def\eg{\emph{e.g., }} 

\def\ie{\emph{i.e., }} 

\def\etc{\emph{etc., }}

\def\year{2021}\relax

\title{Simple is not Easy: A Simple Strong Baseline for TextVQA and TextCaps}
\author{
    Qi Zhu,\textsuperscript{\rm 1} Chenyu Gao,\textsuperscript{\rm 1,2,3} Peng Wang\footnote{Peng Wang is corresponding author.},\textsuperscript{\rm 1 3} Qi Wu\textsuperscript{\rm 4}
}

\affiliations{
    \\
    \textsuperscript{\rm 1} School of Computer Science, Northwestern Polytechnical University, Xi'an, China \\
    \textsuperscript{\rm 2} School of Software, Northwestern Polytechnical University, Xi'an, China \\
    \textsuperscript{\rm 3} National Engineering Laboratory for Integrated Aero-Space-Ground-Ocean Big Data Application Technology, China \\
    \textsuperscript{\rm 4} University of Adelaide, Australia \\
    \{zhu$\_$qi$\_$happy$\_$,chenyugao\}@mail.nwpu.edu.cn, peng.wang@nwpu.edu.cn, qi.wu01@adelaide.edu.au
    
}

\begin{document}

\maketitle

\begin{abstract}
Texts appearing in daily scenes that can be recognized by OCR (Optical Character Recognition) tools contain significant information, such as street name, product brand and prices. Two tasks -- text-based visual question answering and text-based image captioning, with a text extension from existing vision-language applications, are catching on rapidly. 
To address these problems, many sophisticated multi-modality encoding frameworks (such as heterogeneous graph structure) are being used. In this paper, we argue that a simple attention mechanism can do the same or even better job without any bells and whistles. Under this mechanism, we simply split OCR token features into separate visual- and linguistic-attention branches, and send them to a popular Transformer decoder to generate answers or captions. Surprisingly, we find this simple baseline model is rather strong -- it consistently outperforms state-of-the-art (SOTA) models on two popular benchmarks, TextVQA and all three tasks of ST-VQA, although these SOTA models use far more complex encoding mechanisms. Transferring it to text-based image captioning, we also surpass the TextCaps Challenge 2020 winner. We wish this work to set the new baseline for this two OCR text related applications and to inspire  new thinking of multi-modality encoder design. Code is available at \url{https://github.com/ZephyrZhuQi/ssbaseline}
\end{abstract}

\section{Introduction}
\begin{figure}[t!]
	\begin{center}
		\includegraphics[width=0.48\textwidth]{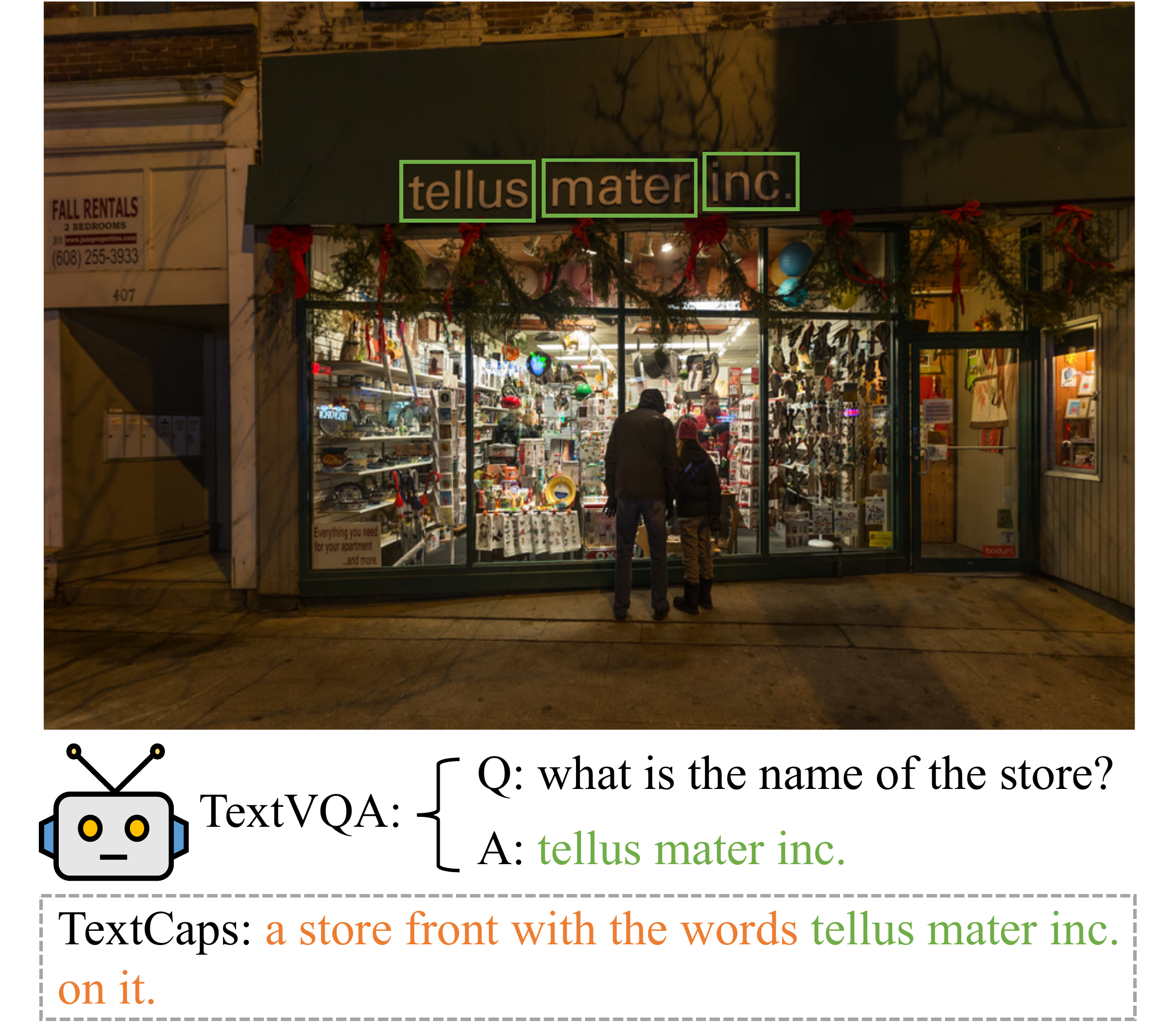}
	\end{center}
	\caption{An example of TextVQA and TextCaps tasks. The answer and description are generated by our model. Our simple baseline is able to read texts and answer related questions. Besides, it can also observe the image and generate a description with texts embedded in it.}
	\label{fig:front}
\end{figure}

To automatically answer a question or generate a description for images that require scene text understanding and reasoning has broad prospects for commercial applications, such as assisted driving and online shopping. Equipped with these abilities, a model can help drivers decide distance to the next street or help customers get more details about a product. Two kinds of tasks that focus on text in images have recently been introduced, which are text-based visual question answering (TextVQA)~\cite{TexVQA, STVQA} and text-based image captioning (TextCaps)~\cite{sidorov2020textcaps}. For example, in Figure~\ref{fig:front}, a model is required to answer a question or generate a description by reading and reasoning the texts ``tellus mater inc.'' in the image. These two tasks pose a challenge to current VQA or image captioning models as they explicitly require understanding of a new modality -- Optical Character Recognition (OCR). A model must efficiently utilize text-related features to solve these problems.

\begin{figure*}[t!]
	\begin{center}
		\includegraphics[width=0.98\textwidth]{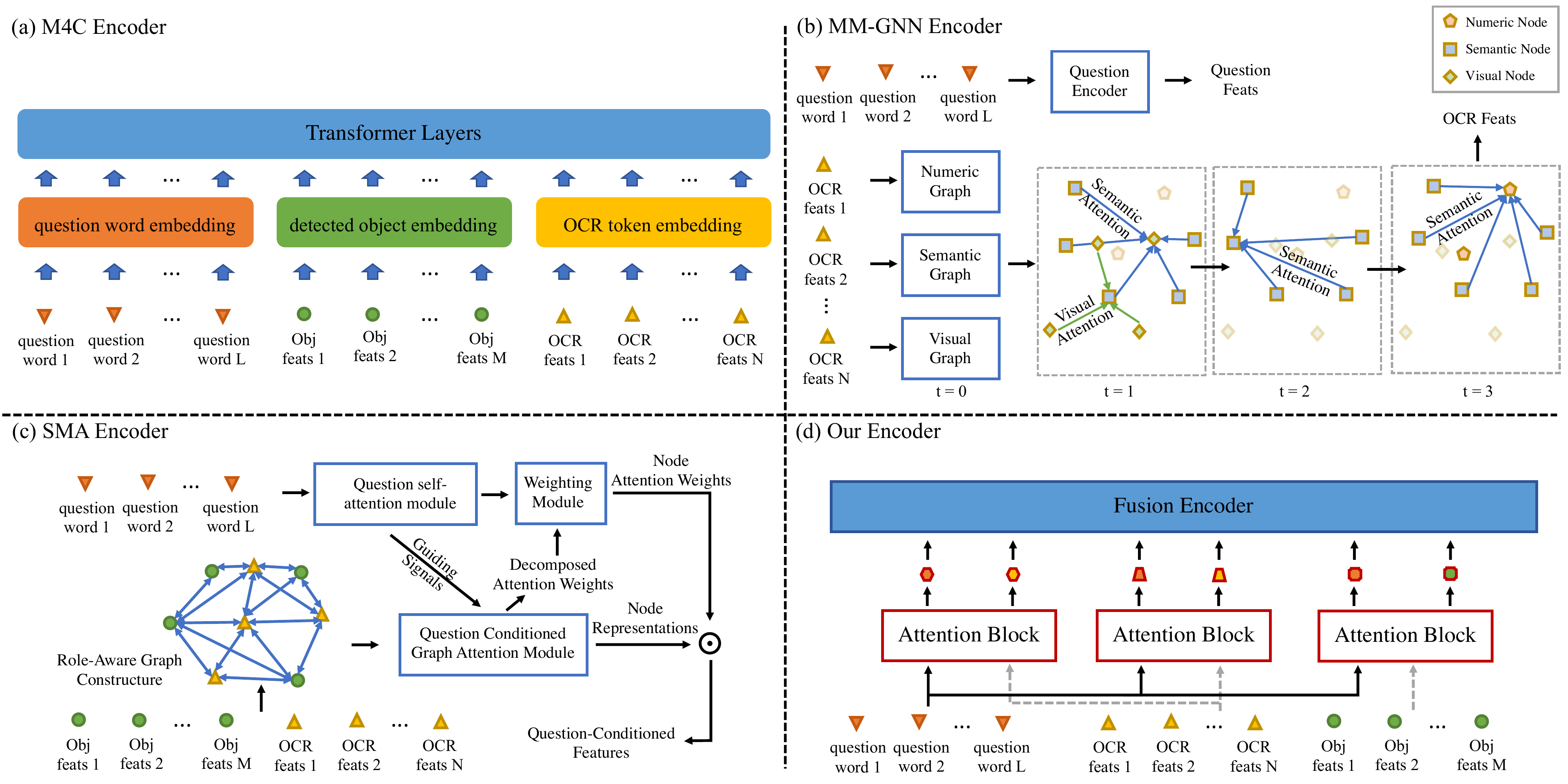}
	\end{center}
	\caption{Encoders of different models. (a) Current state-of-the-art model M4C on TextVQA task forwards each feature vector of all modalities indiscriminately into transformer layers, which exhaust tremendous computation. (b) MM-GNN handcrafts three graphs to represent the image and applies three aggregators step by step to pass messages between graphs. (c) SMA introduces a heterogeneous graph and considers object-object, object-text and text-text relationships, upon which a graph attention network is then used to reason over them. (d) Our baseline uses three vanilla attention blocks to highlight most relevant features and combines them into six individually-functioned vectors, which is then sent into transformer-based fusion encoders. The considerably fewer parameters of six vectors save computation.}
	\label{fig:cmp_encoder}
\end{figure*}

For TextVQA task, the current state-of-the-art model M4C~\cite{hu2019iterative} handles all modalities (questions, visual objects and OCR tokens) over a joint embedding space. Although this homogeneous method is easy to implement, fast to train and has made great headway, it considers that texts and visual objects contribute indiscriminately to this problem and uses text features as a whole. For TextCaps problem, the only difference is that it only has two modalities: visual objects and OCR tokens. However, these limitations remain. 

Some other works proposed even more complex structures to encode and fuse multi-modality features of this task, \ie questions, OCR tokens and images. 
For example, SMA~\cite{gao2020structured} uses a heterogeneous graph to encode object-object, object-text and text-text relationships in the image, and then designs a graph attention network to reason over it. MM-GNN~\cite{Gao_2020_CVPR} represents an image as three graphs and introduces three aggregators to guide message passing from one graph to another.

In this paper, we use the vanilla attention mechanism to fuse pairwise modalities. Under this mechanism, we further use a more reasonable method to utilize text features which leads to a higher performance, that is by splitting text features into two functionally different parts, \ie linguistic- and visual-part which flows into corresponding attention branch. The encoded features are then sent to a popularly-used Transformer-based decoder to generate answers or captions. As compared to the aforementioned M4C models (shown in Figure~\ref{fig:cmp_encoder}a) that throw each instance of every modality into transformer layers, our model (in Figure~\ref{fig:cmp_encoder}d) first uses three attention blocks to filter out irrelevant or redundant features and aggregate them into six individually-functioned vectors. In contrast to hundreds of feature vectors in M4C, the six vectors consume much less computation. Moreover, to group text features into visual- and linguistic- parts is more reasonable. When comparing with the graph-based multi-modal encoders such as MM-GNN (Figure~\ref{fig:cmp_encoder}b) and SMA (Figure~\ref{fig:cmp_encoder}c), our baseline is extremely simple in design and reduces much space and time complexity.

\begin{figure*}[t!]
	\begin{center}
		\includegraphics[width=0.98\textwidth]{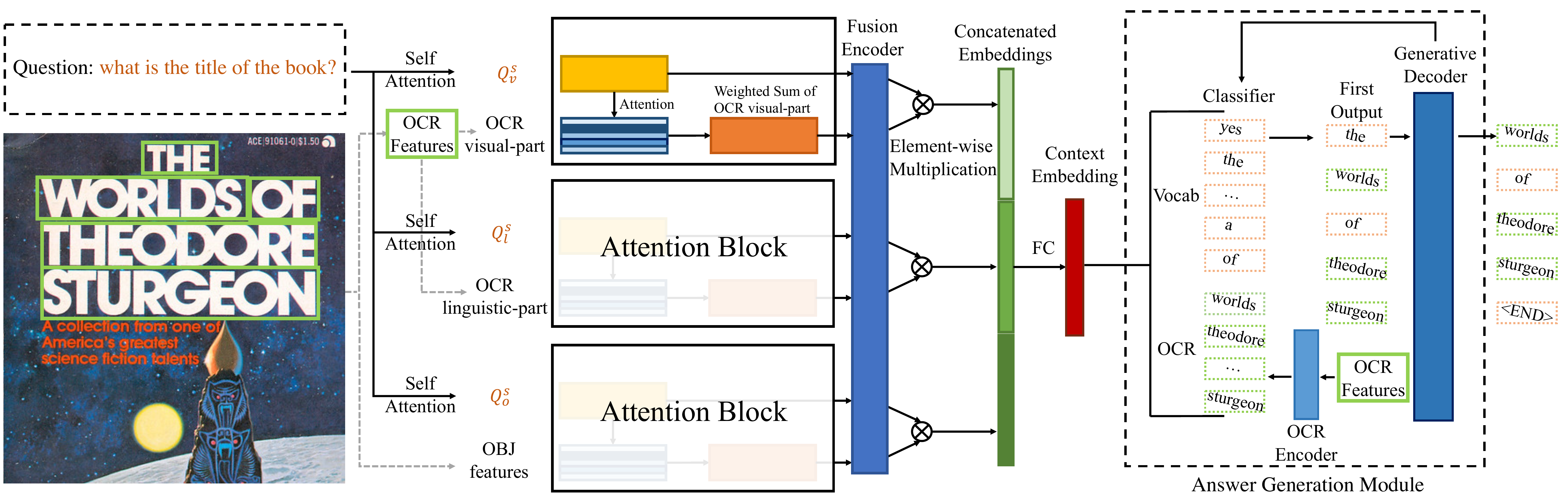}
	\end{center}
	\caption{A simple baseline model for TextVQA. Given an image and a question, we prepare three features (OCR visual-part, OCR linguistic-part and object features) and three question self-attention outputs. The six sequences are put into attention block and fused into six vectors, upon which we calculate element-wise product two by two to get concatenated embeddings. The encoder outputs predict the first word and the rest of answer is produced by an iterative decoder.}
	\label{fig:textvqa_baseline}
\end{figure*}

In addition, for the first time we ask the question: to what extent OCRs contribute to the final performance of TextVQA, in contrast to the other modality - visual contents such as objects and scenes? An interesting phenomenon is observed that OCRs play an almost key role in this special problem while visual contents only serve as assisting factors. A strong model without the use of visual contents surpasses current state-of-the-art model, demonstrating the power of proposed pairwise fusion mechanism and primary role of texts.

To demonstrate the effectiveness of our proposed simple baseline model, we test it on both TextVQA~\cite{TexVQA} and TextCaps~\cite{sidorov2020textcaps} tasks. For the TextVQA, we outperforms the state-of-the-art (SOTA) on TextVQA dataset and all three tasks of  ST-VQA, and rank the first on both leaderboards\footnote{\url{https://eval.ai/web/challenges/challenge-page/551/leaderboard/1575}}\footnote{\url{https://rrc.cvc.uab.es/?ch=11\&com=evaluation\&task=1}}. More importantly, all compared SOTA models use the similar Transformer decoder with ours, but with far more complex encoding mechanisms. For TextCaps, we surpass the TextCaps Challenge 2020 winner and now rank the first place on the leaderboard\footnote{\url{https://eval.ai/web/challenges/challenge-page/573/leaderboard/1617}}.

Overall, the major contribution of this work is to provide a simple but rather strong baseline for the text-based vision-and-language research. This could be the new baseline (backbone) model for both TextVQA and TextCaps. More importantly, we wish this work to inspire a new thinking of multi-modality encoder design -- simple is not easy.


\section{Related Work}
\noindent{\bf Text based visual question answering.}
VQA~\cite{antol2015vqa,johnson2017clevr,FigureQA,FVQA} has seen rapid development in recent years.
A new task -- TextVQA goes one step further and aims at the understanding and reasoning of scene texts in images. A model needs to read texts first and then answer related questions in natural everyday situations.
Two datasets, TextVQA~\cite{TexVQA} and ST-VQA~\cite{STVQA} are introduced concurrently to benchmark progress in this field.
To solve this problem, various methods have also been proposed.
LoRRA, the baseline model in TextVQA, uses bottom-up and top-down~\cite{bottomuptopdown} attention on visual objects and texts to select an answer from either vocabulary or fixed-index OCR.
M4C~\cite{hu2019iterative} is equipped with a vanilla transformer decoder to iteratively generate answers and a flexible pointer network to point back at most probable OCR token at one decoding step.
MM-GNN~\cite{Gao_2020_CVPR} designs a representation of three graphs and introduces three aggregators to update message passing for question answering.

\noindent{\bf Text based image captioning.}
Image captioning challenges a model to automatically generate a natural language description based on the contents in an image. 
Existing datasets, \eg COCO Captions~\cite{chen2015microsoft} and Flickr30k~\cite{young-etal-2014-image}, focus more on visual objects.
To enhance text comprehension in the context of an image, a new dataset called TextCaps \cite{sidorov2020textcaps} is proposed. It requires a model to read and reason about texts and generate coherent descriptions.
The baseline model in TextCaps is modified from aforementioned M4C slightly, by removing question input directly.


\noindent{\bf Generative transformer decoder.}
To address the problem that answers in these two text-based tasks are usually concatenated by more than one word, we use the structure of transformer~\cite{BERT} decoder in answer module.
Following previous work, we also use the generative transformer decoder for fair comparison.


\section{Proposed Method}

Given the three modalities (questions, OCR tokens, visual objects), the first step of our model is to prepare the features by projecting them into the same dimension. Then we describe the formulation of \textit{Attention Block} for feature summarizing. Stacking the blocks together yields encoder for downstream tasks. Using encoder output to produce the first word in answer, we then use an iterative decoder to predict the rest of words. This whole process is shown in Figure~\ref{fig:textvqa_baseline}. When transferring to TextCaps, we only make minimal modifications which will be detailed in Section~\ref{sec:stack_block}.

\noindent {\bf Notation} In the remainder of this paper, all $\bW$ are learned linear transformations, with different symbols to denote independent parameters, \eg $\bW_{fr}$.
$\mathrm{LN}$ is Layer Normalization~\cite{ba2016layer}. 
$\circ$ represents element-wise product. 

\subsection{Feature Preparation}

\noindent{\bf Question features.}
For a question with $L$ words, we first use a three-layer BERT~\cite{BERT} model to embed it into $Q = \{q_{i}\}_{i=1}^{L}$. This BERT model is finetuned during training. 

\noindent{\bf OCR features.}
In text-based VQA and image captioning problem, texts are of key importance. Simply gather every feature of text together is not efficient enough. When faced with a bunch of OCRs, human recognition system tends to use two complementary methods to select subsequent words, either by finding similarly-looking and spatially-close words or choosing words coherent in linguistic meaning. For this intuitive purpose, we split features of $N$ OCR tokens into two parts: \textit{Visual} and \textit{Linguistic}.

\noindent{\textit{1) OCR visual-part. }}
Visual features are combined by appearance feature and spatial feature as they show what eyes catch of, without further processing of natural language system. From this part, a model can get visual information such as word font, color and background. These two are extracted by an off-the-shelf Faster R-CNN~\cite{ren2015faster} detector.

\begin{equation}
{
	\begin{split}
	    & \mathbf{x}_{i}^{ocr,v} = \mathrm{LN}( \bW_{fr}\mathbf{x}_{i}^{ocr,fr})
	    + \mathrm{LN}(\bW_{bx}\mathbf{x}_{i}^{ocr,bx}),
	\end{split}
	\label{ocr_visual}}
\end{equation}
where $\mathbf{x}_{i}^{ocr,fr}$ is the appearance feature extracted from the fc6 layer of Faster R-CNN detector. The fc7 weights are finetuned on our task. $\mathbf{x}_{i}^{ocr,bx}$ is the bounding box feature in the format of $[x_{tl},y_{tl},x_{br},y_{br}]$, where $tl$ and $br$ denotes top left and bottom right coordinates respectively.

\noindent{\textit{2) OCR linguistic-part. }}
Linguistic features are made up of 1) FastText feature $\mathbf{x}_{i}^{ocr,ft}$, which is extracted from a pretrained word embedding and 2) character-level Pyramidal Histogram of Characters (PHOC)~\cite{almazan2014word} feature $x_{i}^{ocr,ph}$ as they contain natural language related information.

\begin{equation}
{
	\begin{split}
	    & \mathbf{x}_{i}^{ocr,l} = \mathrm{LN}( \bW_{ft}\mathbf{x}_{i}^{ocr,ft} +  \bW_{ph}\mathbf{x}_{i}^{ocr,ph})
	\end{split}
	\label{ocr_linguistic}}
\end{equation}

\noindent{\textit{3) OCR additional features. }}In the SBD-Trans~\cite{SBD,textTrans} that we use to recognize OCR tokens, the holistic representations in a specific text region are themselves visual features, however, are employed for linguistic word classification purpose. Therefore, they cover both visual and linguistic context of the OCR token and we thus introduce the Recog-CNN feature $\mathbf{x}_{i}^{ocr,rg}$ from this network to enrich text features.

Finally, Recog-CNN features are added to OCR visual- and linguistic-part simultaneously.

\begin{equation}
{
	\begin{split}
    	& \mathbf{x}_{i}^{ocr,v} = \mathrm{LN}( \bW_{fr}\mathbf{x}_{i}^{ocr,fr} + \bW_{rg}\mathbf{x}_{i}^{ocr,rg})
	    + \mathrm{LN}(\bW_{bx}\mathbf{x}_{i}^{ocr,bx})\\
	    & \mathbf{x}_{i}^{ocr,s} = \mathrm{LN}( \bW_{ft}\mathbf{x}_{i}^{ocr,ft} +  \bW_{ph}\mathbf{x}_{i}^{ocr,ph} + \bW_{rg}\mathbf{x}_{i}^{ocr,rg})
	\end{split}
	\label{ocr_trans}}
\end{equation}

\noindent{\bf Visual features.}
In text-based tasks, visual contents in an image can be utilized to assist textual information in the reasoning process. To prove that our simple attention block has the power of using visual features in various forms, we adopt either grid-based global features or region-based object features.

\noindent{\textit{1) Global features. }}
We obtain image global features $x_{i}^{glob}$ from a ResNet-152~\cite{he2016deep} model pretrained on ImageNet, by average pooling $2048D$ features from the res-5c block, yielding a $14 \times 14 \times 2048$ feature for one image. To be consistent with other features, we resize the feature into $196 \times 2048$, a total of $196$ uniformly-cut grids.

\begin{equation}
{
	\begin{split}
	    & \mathbf{x}_{i}^{glob} = \mathrm{LN}( \bW_{g}\mathbf{x}_{i}^{glob})
	\end{split}
	\label{obj_glob}}
\end{equation}

\noindent{\textit{2) Object features. }}
The region-based object features are extracted from the same Faster F-CNN model as mentioned in OCR features part.

\begin{equation}
{
	\begin{split}
	    & \mathbf{x}_{i}^{obj} = \mathrm{LN}( \bW_{fr}^{'}\mathbf{x}_{i}^{obj,fr})
	    + \mathrm{LN}(\bW_{bx}^{'}\mathbf{x}_{i}^{obj,bx}),
	\end{split}
	\label{obj_visual}}
\end{equation}
where $\mathbf{x}_{i}^{obj,fr}$ is the appearance feature and $\mathbf{x}_{i}^{obj,bx}$ is the bounding box feature.

\subsection{Attention Block as Feature Summarizing}
\label{sec:attn_block}
In tasks that cross the fields of computer vision and natural language processing, modality fusion is of superior importance. Treating them as homogeneous entities in a joint embedding space might be easy to implement, however, is not carefully tailored to a specific problem. Moreover, the many parameters of all entities in the large model (for example, a Transformer) consume much computation.
To grasp interaction between modalities for maximum benefit and filter out irrelevant or redundant features before the entering into a large fusion layer, we use a simple attention block to input two sequences of entities and output two processed vectors, which is shown as the \textit{Attention Block} in Figure~\ref{fig:textvqa_baseline}.

The two sequences of entities might be any sequence we want. For TextVQA problem, question changes in real-time and plays a dominant role in final answering. The design of question needs careful consideration and its existence should contribute throughout the process. For example, here we use question as one input of \textit{query} in attention block. The sequence of question words goes through a self-attention process before forwarding into attention block.

First we put the question word sequence $Q = \{q_{i}\}_{i=1}^{L}$ through a fully connected feed-forward network, which consists of two linear transformations (or two convolutions with $1$ as kernel size) and one ReLU activation function between them.
\begin{equation}
	\begin{split}
	&q_{i}^{fc} = \mathrm{conv}\{\mathrm{ReLU}[\mathrm{conv}(q_{i})]\}, \,\, i = 1, \dots, L ; 
	\end{split}
\end{equation}
A softmax layer is the used to compute the attention on each word in the question.
\begin{equation}
	\begin{split}
	&a_i = \mathrm{Softmax}({q_{i}^{fc}}), \,\, i = 1, \dots, L ;\\
	\end{split}
\end{equation}
This is known as self-attention and these weights are multiplied with original question embedding to get weighted sum of word embeddings.
\begin{equation}
	\begin{split}
	&Q^s = \textstyle{\sum_{i=1}^{L}} a_i  {q}_{i}.
	\end{split}
\end{equation}
If we have several individual entities to combine with question, the corresponding number of parallel self-attention processes are performed on the same question with independent parameters. For example, we can get $Q^s_{v}$, $Q^s_{l}$ and $Q^s_{o}$ for OCR visual-part, OCR linguistic-part and object regions respectively.

Then the $Q^s$ are used as query for corresponding features. We calculate the attention weights under the guidance of $Q^s$, which are then put into a softmax layer. Finally the weights are multiplied with original queried features to get a filtered vector. Here we take the pair of $Q^s_{v}$ and ${\mathbf{x}}_{i}^{ocr,v}$ as an example:
\begin{equation}
{
	\begin{split}
		&p_i = \bW[\mathrm{ReLU}(\bW_{s} Q^s_{v})\circ \mathrm{ReLU}(\bW_{x}{x}_i^{ocr,v})],\\
		&s_i = \mathrm{Softmax}(p_i), \quad  i=1,\dots,N,\\
		&g^{ocr,v} = \textstyle{\sum_{i=1}^{N}} s_i  {\mathbf{x}}_{i}^{ocr,v}
	\end{split}
	\label{attn}}
\end{equation}
where $g^{ocr,v}$ is the output of attention block. Similarly, we can get $g^{ocr,l}$ for the OCR-linguistic summarizing feature, $g^{obj}$ for the object summarizing feature. 

{Different from M4C sending every single question tokens, OCR tokens and objects into the transformer feature fusion layer, here we only have 6 feature vectors ($Q^s_{v}$, $Q^s_{l}$, $Q^s_{o}$, $g^{ocr,v}$, $g^{ocr,l}$ and $g^{obj}$) which are sent to the following process. This largely decreases the computation complexity and burden, considering that the transformer is a parameter-heavy network.} 

\subsection{Stacked-Block Encoder}
\label{sec:stack_block}
The attention block in Section~\ref{sec:attn_block} can be stacked together as an encoder which produces combined embedding for downstream tasks.

\noindent{\bf TextVQA baseline model. }
As presented in the above module, questions are sent through self-attention to output $Q^s_{v}$, $Q^s_{l}$ and $Q^s_{o}$. OCR features in images are splitted into visual- and linguistic part, which are $x^{ocr,v}$ and $x^{ocr,l}$. We also have object features $x^{obj}$.
The six sequences are put into three attention blocks and we get six $768D$ vectors which are then forwarded into a fusion encoder. The fusion encoder, OCR encoder and generative decoder in Figure~\ref{fig:textvqa_baseline} are in the same transformer model but undertaking different roles.
After fusion encoder processing, the six vectors conduct element-wise multiplication in a pairwise way to get corresponding embeddings which are concatenated together. Then we use a fully-connected layer to transform the concatenated embeddings to a context embedding with appropriate dimension, upon which we generate the first answer output. Given the first answer word, a generative decoder is then used to select the rest of answer, which will be detailed in Section~\ref{sec:ans_mod} and Supplementary Material.


\begin{figure}[t!]
	\begin{center}
		\includegraphics[width=0.48\textwidth]{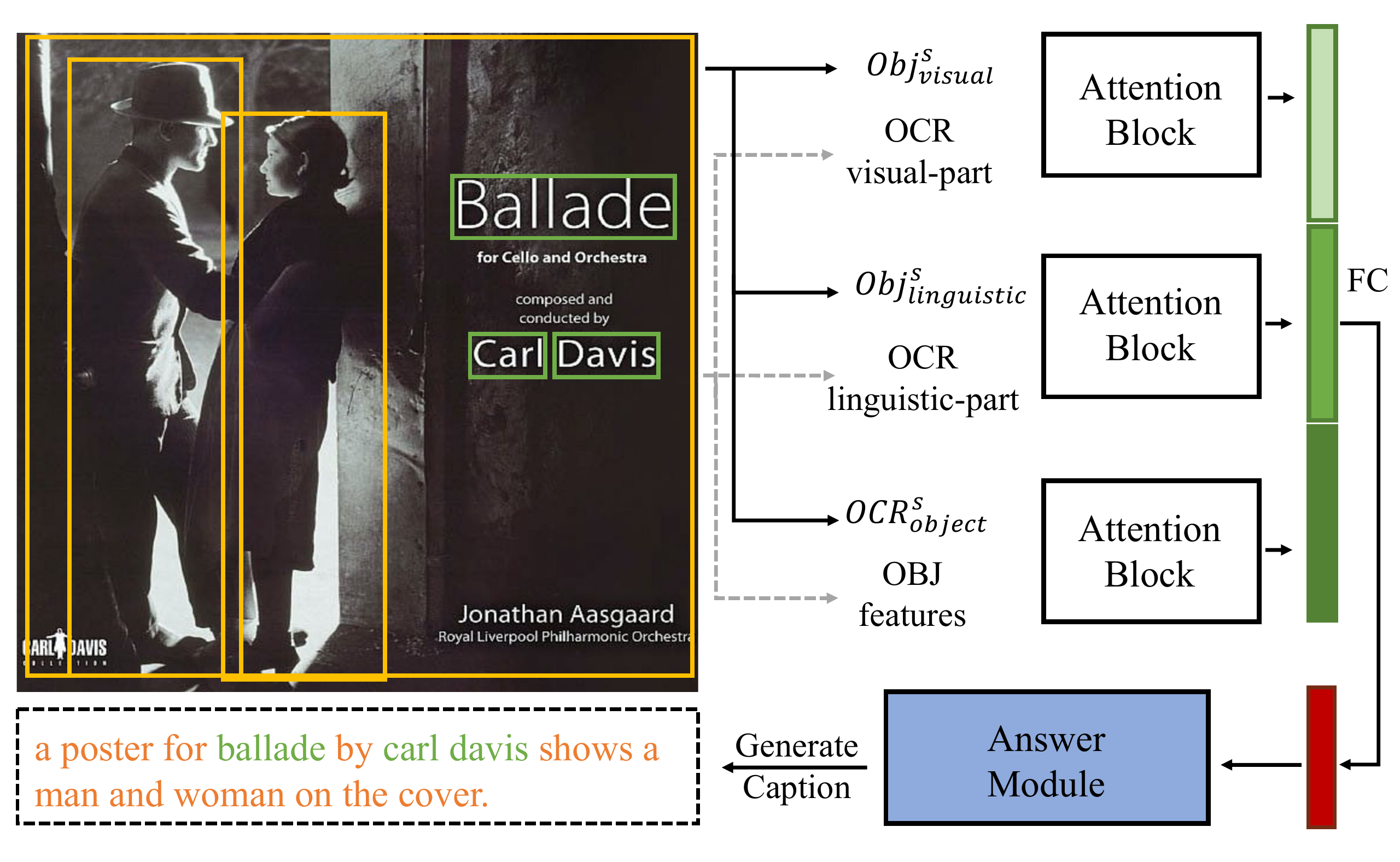}
	\end{center}
	\caption{TextCaps baseline model. It has a same structure as TextVQA baseline model.}
	\label{fig:textcaps_baseline}
\end{figure}

\begin{center}
    \begin{table*}[t!]
    \centering
    \scalebox{0.95}{
    \begin{tabular}{ccccclcc}
    \toprule[1pt]
    \# & Method               & \begin{tabular}[c]{@{}c@{}}Trans\\ structure\end{tabular} & \begin{tabular}[c]{@{}c@{}}OCR \\ system\end{tabular} & \begin{tabular}[c]{@{}c@{}}Visual\\ feat\end{tabular} & \multicolumn{1}{c}{\begin{tabular}[c]{@{}c@{}}OCR \\ feat\end{tabular}} & \begin{tabular}[c]{@{}c@{}}Accu. \\ on val\end{tabular} & \begin{tabular}[c]{@{}c@{}}Accu.\\ on test\end{tabular} \\ \hline \hline
    1  & one-block & 4-layer                                 & Rosetta-en & - & Fast, PHOC, FRCN, bbox & $37.51$ & - \\ 
    2  & two-block & 4-layer & Rosetta-en & - & Fast, PHOC, FRCN, bbox & $39.28$ & $39.99$ \\ 
    3  & three-block & 4-layer & Rosetta-en  & Global & Fast, PHOC, FRCN, bbox & $39.52$ & - \\ 
    4  & three-block & 4-layer & Rosetta-en & Obj & Fast, PHOC, FRCN, bbox & $39.91$ & - \\ 
    5  & three-block & 4-layer & Rosetta-en & Obj & Fast, PHOC, FRCN, bbox, Recog-CNN & $40.28$ & 
    - \\
    6  & three-block & 8-layer & Rosetta-en & Obj & Fast, PHOC, FRCN, bbox, Recog-CNN & $40.38$ & $40.92$ \\ 
    7  & three-block & 8-layer & SBD-Trans & Obj & Fast, PHOC, FRCN, bbox, Recog-CNN & $43.95$ & $44.72$ \\ 
    8  & three-block(w/ST-VQA) & 8-layer & SBD-Trans & Obj & Fast, PHOC, FRCN, bbox, Recog-CNN & $\textbf{45.53}$ & $\textbf{45.66}$ \\ \bottomrule[1pt]
    \end{tabular}
    }
    \caption{We ablate our model on TextVQA dataset by testing number of attention blocks, forms of visual object features and addition of OCR representations.}
    \label{tab:ablation_textvqa_}
    \end{table*}
\end{center}

\begin{center}
    \begin{table}[htbp]
    \centering
    \scalebox{0.88}{
	\setlength{\tabcolsep}{1mm}{
    \begin{tabular}{clccc}
    \toprule[1pt]
    \multicolumn{1}{c}{\#} & \multicolumn{1}{c}{Method} & \multicolumn{1}{c}{\begin{tabular}[c]{@{}c@{}}OCR \\ system\end{tabular}} & \multicolumn{1}{c}{\begin{tabular}[c]{@{}c@{}}Accu.\\ on val\end{tabular}} & \multicolumn{1}{c}{\begin{tabular}[c]{@{}c@{}}Accu.\\ on test\end{tabular}} \\ \hline\hline
    $1$ & LoRRA~\cite{TexVQA}  & Rosetta-ml & $26.56$   & $27.63$     \\
    $2$ & DCD ZJU~\cite{DCDZJU}   & - & $31.48$   & $31.44$ \\
    $3$ & MSFT VTI~\cite{MSFTVTI}   & -  & $32.92$   & $32.46$     \\
    $4$ & M4C~\cite{hu2019iterative} & Rosetta-en & $39.40$ & $39.01$ \\ 
    $5$ & SA-M4C~\cite{sam4c} & Google OCR & $45.40$ & $44.60$ \\
    $6$ & SMA~\cite{SMA} & SBD-Trans & $44.58$ & $45.51$ \\
    $7$ & ours (three-block) & Rosetta-en & $40.38$ & $40.92$ \\ 
    $8$ & ours (three-block w/ST-VQA) & SBD-Trans & $\textbf{45.53}$ & $\textbf{45.66}$ \\ 
    \bottomrule[1pt]
    \end{tabular}
    }
	}
    \caption{Comparison to previous work on TextVQA dataset. Our model sets new state-of-the-art with an extremely simple design.}
    \label{tab:cmp_textvqa}
    \end{table}
\end{center}
\noindent{\bf TextCaps baseline model. }
As there are no questions in TextCaps, we use objects to guide OCR visual- and linguistic-part and use OCRs to guide object features. Technically we simply replace question word sequence with OCR token sequence or object proposal feature sequence. The other settings are the same with TextVQA. Figure~\ref{fig:textcaps_baseline} illustrates our Textcaps baseline model.
To easily transfer to another task demonstrates the generalization ability and simplicity of our method.

\subsection{Answer Generation Module}
\label{sec:ans_mod}
To answer a question or generate a caption, we use a generative decoder based on transformer. 
It takes as input the `context embedding' from the above encoder and select the first word of the answer. Based on the first output word, we then use the decoder to find the next word token either from a pre-built vocabulary or the candidate OCR tokens extracted from the given image, based on a scoring function.


\noindent{\bf Training Loss.}
\label{sec:training_loss}
Considering that the answer may come from two sources, we use multi-label binary cross-entropy (bce) loss:
\begin{gather}
{
    \begin{split}
    pred &= \frac{1}{1+\exp{(-y_{pred})}}, \\
    \mathcal{L}_{bce} &= -y_{gt} \mathrm{log}(pred)-(1-y_{gt}) \mathrm{log}(1-pred),
    \end{split}
}
\end{gather}
where $y_{pred}$ is prediction and $y_{gt}$ is ground-truth target.

\noindent{\bf Additional Training Loss.}
In some cases, the model reasons correctly, however, picks slightly different words than what we expected due to defective reading (OCR) ability. 
To take advantage of these predictions, we introduce a new policy gradient loss as an auxilliary task inspired by reinforcement learning. In this task, the greater reward, the better. We take Average Normalized Levenshtein Similarity (ANLS) metric\footnote{$\mathrm{ANLS}(s_{1},s_{2})=1-d(s_{1},s_{2})/\mathrm{max}(\mathrm{len}(s_{1}),\mathrm{len}(s_{2}))$, d(,) is edit distance.} as the reward which measures the character similarity between predicted answer and ground-truth answer.

\begin{equation}
	\begin{split}
    &r = \mathrm{ANLS}(\phi(y_{gt}),\phi(y_{pred})), \\
    &y = \mathbb{I}(\mathrm{softmax}(y_{pred})), \\
    &\mathcal{L}_{pg} = (0.5 - r) (y_{gt} \mathrm{log}(y) + (1-y_{gt}) \mathrm{log}(1-y)),\\
    &\mathcal{L} = \mathcal{L}_{bce}+\alpha\cdot\mathcal{L}_{pg},
    \end{split}
\end{equation}
where $\phi$ is a mapping function that returns a sentence given predicting 
score (\eg, $y_{pred}$), ANLS($\cdot$) is used to calculate similarity between two phrases, $\mathbb{I}$ is an indicator function to choose the maximum probability element. The additional training loss is a weighted sum of $\mathcal{L}_{bce}$ and $\mathcal{L}_{pg}$, where $\alpha$ is a hyper-parameter to control the trade-off of $\mathcal{L}_{pg}$.
After introducing policy gradient loss, our model is able to learn fine-grained character composition alongside linguistic information. We only apply this additional loss on ST-VQA dataset, which brings roughly $1\%$ improvement.

\section{Experiments}
Extensive experiments are conducted across two categories of tasks: TextVQA and TextCap. For TextVQA we set new state-of-the-art on TextVQA dataset and all three tasks of ST-VQA. For TextCaps we surpass 2020 TextCaps Challenge winner. See more experiments details below.

\subsection{Implementation Details}
The set of methods are built on top of PyTorch. We use Adam as the optimizer. The learning rate for TextVQA and TexCaps is set to $1e-4$. For TextVQA we multiply the learning rate with a factor of $0.1$ at the $14,000$ and $15,000$ iterations in a total of $24,000$ iterations. For TextCaps the multiplication is done at the $3,000$ and $4,000$ iterations, with $12,000$ total iterations.
We set the maximum length of questions to $L = 20$. We recognize at most $N = 50$ OCR tokens and detect at most $M = 100$ objects. The maximum number of decoding steps is set to $12$. Transformer layer in our model uses $12$ attention heads. The other hyper-parameters are the same with BERT-BASE~\cite{BERT}. We use the same model on TextVQA and three tasks of ST-VQA, only with different answer vocabulary, both with a fixed size of $5000$.

\subsection{Ablation Study on TextVQA Dataset}
TextVQA~\cite{TexVQA} is a popular benchmark dataset to test scene text understanding and reasoning ability, which contains $45,336$ questions on $28,408$ images. Diverse questions involving inquires about time, names, brands, authors, \etc and dynamic OCR tokens that might be rotated, casual or partially occluded make it a challenging task.

We first conduct an experiment of building only one block, with one question self-attention output guiding the whole set of text features as a comparison. This is a \textbf{one-block} model in Table\ref{tab:ablation_textvqa_} which does not perform as good as the state-of-the-art. To investigate how well text features can perform without the usage of visual grid-based or region-based features, we build a \textbf{two-block} model. Given the two categories of OCR features -- visual and linguistic, we find that our simple model is already able to perform promisingly on TextVQA problem. Line $1$ and Line $2$ tell clearly the validity of sorting text features into two groups ($1.77\%$ difference). 

When building our third block on the basis of visual contents, either global features or object-level features are at our disposal. The incorporation of a third block has modest improvements ($0.24\%$ for global and $0.63\%$ for object features).  
From Line $4$ to Line $5$, a new Recog-CNN feature is added to enrich text representation and brings $0.37\%$ improvement. We also use more transformer layers (from $4$ to $8$) and get $0.1\%$ higher result. 
Then we use a much better OCR system (especially on recognition part) and obtain large performance boost (from $40.38\%$ to $43.95\%$).

\begin{center}
	\begin{table}[t]
	\centering
	\scalebox{0.95}{
	\setlength{\tabcolsep}{1mm}{
		\begin{tabular}{lllll}
        \toprule[1pt]
        \multicolumn{1}{c}{\#} & \multicolumn{1}{c}{Method} & \multicolumn{1}{c}{\begin{tabular}[c]{@{}c@{}}ANLS \\ on test1\end{tabular}} & \multicolumn{1}{c}{\begin{tabular}[c]{@{}c@{}}ANLS\\ on test2\end{tabular}} & \multicolumn{1}{c}{\begin{tabular}[c]{@{}c@{}}ANLS\\ on test3\end{tabular}}\\ \hline \hline
        1 &M4C~\cite{hu2019iterative} & - & - & $0.4621$\\
        2 &SA-M4C~\cite{sam4c} & - & $0.4972$ & $0.5042$\\ 
        3 &SMA~\cite{SMA} & $0.5081$ & $0.3104$ & $0.4659$\\ 
        4 &ours & $0.5060$ & $0.5047$ &$0.5089$ \\
        5 &ours(w/TextVQA) & $\textbf{0.5490}$ & $\textbf{0.5513}$ & $\textbf{0.5500}$ \\
        \bottomrule[1pt]
        \end{tabular}
	}
	}
	\caption{\label{tab:ablation_stvqa}Comparison to previous work on ST-VQA dataset. With TextVQA pretraining, our model outperforms current approaches by a large margin.
	}
	\end{table}
\end{center}

\begin{figure*}[t!]
	\begin{center}
		\includegraphics[width=0.98\textwidth]{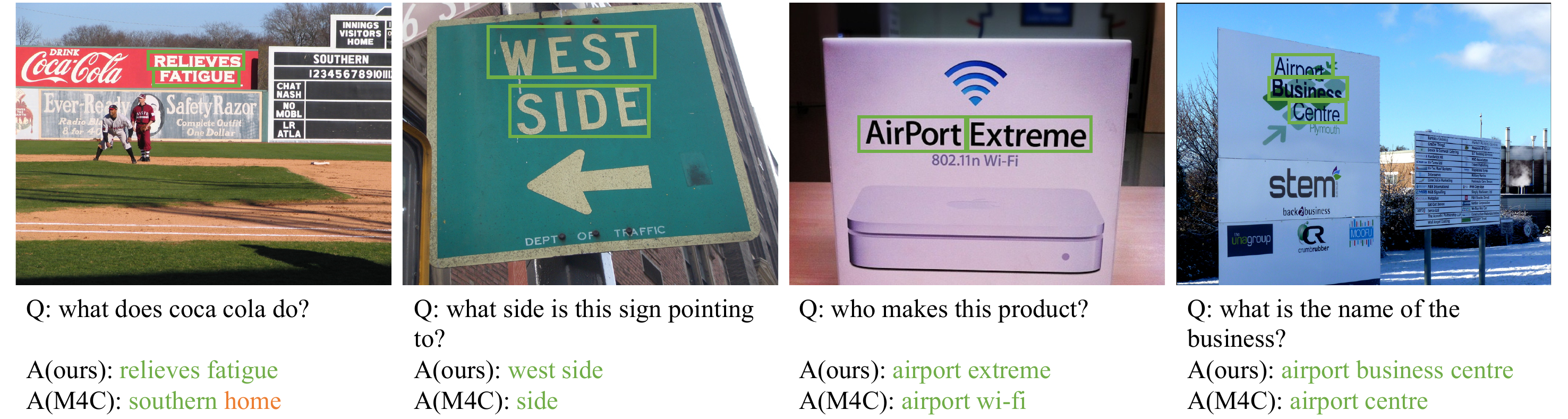}
	\end{center}
	\caption{Qualitative examples of our baseline model in contrast to M4C. While our model can read texts in accordance with written language system, M4C can only pick tokens in a random and errorneous way. }
	\label{fig:suc_case}
\end{figure*}

\noindent{\bf Qualitative examples. } We present four examples in Figure~\ref{fig:suc_case} where our model shows the ability to really read scene texts in a way similar to humans, \ie left-to-right then top-to-bottom. In contrast, current state-of-the-art model M4C fails to read tokens in a correct order. 

\subsection{Comparison with state-of-the-art}
\noindent{\bf TextVQA dataset. }
Even stripped of the usage of visual contents in the image, our two-block model already surpasses current state-of-the-art M4C by $0.98\%$ on test set (Line $2$ in Table~\ref{tab:ablation_textvqa_} VS. Line $4$ in Table~\ref{tab:cmp_textvqa}). Using the same OCR system, our baseline model further improves upon M4C by $0.98\%$ on val and $1.91\%$ on test (Line $7$ VS. Line $4$ in Table~\ref{tab:cmp_textvqa}).

Compared to the top entries in TextVQA Challenge 2020, our baseline has a significantly simpler model design, especially on the encoder side. M4C and SA-M4C take all parameters of entities into transformer layers and join large amout of computation. SMA uses a heterogeneous graph to explicitly consider different nodes and compute attention weights on $5$-neighbored graph. Our model that surpasses all of them only sends six holistic vectors two-by-two into transformer layers, which tremendously saves computation.

\noindent{\bf ST-VQA dataset. } The ST-VQA dataset~\cite{STVQA} is another popular dataset with three tasks, which gradually increase in difficulty. Task 1 provides a dynamic candidate dictionary of $100$ words per image, while Task 2 provides a fixed answer dictionary of $30,000$ words for the whole dataset. As for Task 3, however, the model are supposed to generate answer without extra information. We also evaluate our model on ST-VQA dataset, using the same model from TextVQA for all three tasks. Without any additional training data, our model achieved the highest on Task 2 and Task 3 (Line $4$ in Table~\ref{tab:ablation_stvqa}). Using TextVQA dataset as additional training data, our model sets new state-of-the-art on all three tasks and outperforms current approaches by a large margin.

\begin{center}
    \begin{table}[t!]
    \centering
    \scalebox{0.98}{
	\setlength{\tabcolsep}{1mm}{
    \begin{tabular}{ccccccc}
    \toprule[1pt]
    \multirow{2}{*}{\#} & \multirow{2}{*}{Method} & \multicolumn{5}{c}{Val set metrics} \\ \cline{3-7} 
     &  & B & M & R & S & C \\ \hline
    1 & M4C-Captioner & $23.30$ & $22.00$ & $46.20$ & $15.60$ & $89.60$ \\
    2 & ours(Rosetta-en) & $23.87$ & $22.02$ & $46.4$ & $15.01$ & $91.06$ \\
    3 & ours(SBD-Trans) & $\textbf{24.89}$ & $\textbf{22.71}$ & $\textbf{47.24}$ & $\textbf{15.71}$ & $\textbf{98.83}$ \\ \hline \hline
    \multirow{2}{*}{\#} & \multirow{2}{*}{Method} & \multicolumn{5}{c}{Test set metrics} \\ \cline{3-7} 
     &  & B & M & R & S & C \\ \hline
    4 & M4C-Captioner & $18.9$ & $19.8$ & $43.2$ & $12.8$ & $81.0$ \\
    5 & colab\_buaa(Winner) & $20.09$ & $\textbf{20.62}$ & $\textbf{44.30}$ & $\textbf{13.50}$ & $88.48$ \\
    6 & ours(SBD-Trans) & $\textbf{20.16}$ & $20.3$ & $44.23$ & $12.82$ & $\textbf{89.63}$ \\
    \bottomrule[1pt]
    \end{tabular}
    }
    }
    \caption{Results on TextCaps dataset. (B: BLEU-4; M: METEOR; R: ROUGE\_L; S: SPICE; C: CIDEr)}
    \label{tab:textcaps}
    \end{table}
\end{center}

\subsection{TextCaps Dataset}
TextCaps is a new dataset that requires a model to read texts in images and generate descriptions based on scene text understanding and reasoning. In TextCaps, automatic captioning metrics (BLEU~\cite{papineni2002bleu}, METEOR~\cite{denkowski2014meteor}, ROUGE\_L~\cite{lin2004rouge}, SPICE~\cite{anderson2016spice} and CIDEr~\cite{vedantam2015cider}) are compared with human evaluation scores.
All automatic metrics show high correlation with human scores, among which CIDEr and METEOR have the highest.

M4C-Captioner is the method provided in TextCaps, which is modified from M4C model by simply removing question input. Similarly, simply replacing question in our TextVQA baseline model with object or OCR sequence yields our TextCaps baseline model. Using exactly the same OCR system, OCR representations, our baseline with Rosetta-en OCR (Line $2$ on Table~\ref{tab:textcaps}) already surpass M4C-Captioner (Line $1$ on Table~\ref{tab:textcaps}), especially on BLUE-4 and CIDEr metric. By upgrading our OCR system to SBD-Trans and using $6$-layer transformer in our encoder-decoder structure, our baseline further exceeds TextCaps Challenge Winner on BLUE-4 and CIDEr metric as shown in Line $5$ and Line $6$ of Table~\ref{tab:textcaps}.


\begin{algorithm*}[h]
    \caption{Answer generation in TextVQA}
    \label{alg:textvqa}
    \KwIn{Question features $Q$, OCR features $x^{ocr}$, Object features $x^{obj}$\;}
    \KwOut{An answer with a length of $T$ words $A=\{a_{i}\}_{i=0}^{T-1}$\;}
    Obtain three independent question signals:\\
    \nonl \hspace{1.3em} $(Q^s_{v}, Q^s_{l}, Q^s_{o}) = \mathrm{Self\_Attention}^{\{v,l,o\}}(Q)$\;
    Split OCR features into OCR visual- and linguistic-part ($x^{ocr,v}$, $x^{ocr,l}$)\;
    Forward question signals and three features into three \textit{Attention Blocks} in a pairwise way:\\
    \nonl \hspace{1.3em} $g^{ocr,v} = \mathrm{Att\_Blk}(Q^s_{v}, x^{ocr,v});\ g^{ocr,l} = \mathrm{Att\_Blk}(Q^s_{l}, x^{ocr,l});\ g^{obj} = \mathrm{Att\_Blk}(Q^s_{o}, x^{obj})$\;
    Forward six vectors into fusion encoder:\\
    \nonl \hspace{1.3em} $\bar{Q}^s_{v}, \bar{Q}^s_{l}, \bar{Q}^s_{o}, \bar{g}^{ocr,v}, \bar{g}^{ocr,l}, \bar{g}^{obj} = \mathrm{Fusion\_Encoder}(Q^s_{v}, Q^s_{l}, Q^s_{o}, g^{ocr,v}, g^{ocr,l}, g^{obj})$\;
    Element-wise multiply the six vectors two by two:\\
    \nonl \hspace{1.3em} $\be^{ocr,v}=\bar{Q}^s_{v}\circ \bar{g}^{ocr,v};\ \be^{ocr,l}=\bar{Q}^s_{l}\circ \bar{g}^{ocr,l};\ \be^{obj}=\bar{Q}^s_{o}\circ \bar{g}^{obj}$\;
    Concatenate the three results together and transform it into \textbf{context embedding}\;
    \nonl \hspace{1.3em} $\mathrm{context\, embedding}=\mathrm{Linear}([\be^{ocr,v}; \be^{ocr,l}; \be^{obj}])$\;
    $t = 0$\;
    Based on \textit{context embedding}, calculate the first word $a_{t}$ in the answer:\\
    \nonl \hspace{1.3em} 1) Map \textit{context embedding} to a $5000D$ vector through a linear layer as scores for vocabulary\;
    \nonl \hspace{1.3em} 2) Get $50D$ dot product between \textit{context embedding} and all $50$ OCR encoder outputs as scores for OCR tokens\;
    \nonl \hspace{1.3em} 3) Concatenate $5000D$ vector and $50D$ vector together, from which select the highest-score index as $a_{t}$\;
    \For{$t=1;t \le T-1;t++$}
    {
      \If{the previous output $a_{t-1}$ is from vocabulary}
      {
        Forward linear layer weights corresponding to the previous output $a_{t-1}$ into generative decoder and obtain \textit{decoder embedding}\;
      }
      \ElseIf{the previous output $a_{t-1}$ is from OCR tokens}
      {
        Forward OCR features of the previous output $a_{t-1}$ into generative decoder and obtain \textit{decoder embedding}\;
      }
      Based on \textit{decoder embedding}, calculated the current word $a_{t}$ in the answer:\\
      \nonl \hspace{1.3em} 1) Map \textit{decoder embedding} to a $5000D$ vector through a linear layer as scores for vocabulary\;
      \nonl \hspace{1.3em} 2) Get $50D$ dot product between \textit{decoder embedding} and all $50$ OCR encoder outputs as scores for OCR tokens\;
      \nonl \hspace{1.3em} 3) Concatenate $5000D$ vector and $50D$ vector together, from which select the highest-score index as $a_{t}$\;
    }
    return $A=\{a_{i}\}_{i=0}^{T-1}$\;
\end{algorithm*}

\begin{center}
    \begin{table*}[t!]
    \centering
    \scalebox{1}{
    \setlength{\tabcolsep}{3mm}{
    \begin{tabular}{ccccc}
    \toprule[1pt]
    \multirow{2}{*}{\#} & \multirow{2}{*}{Model} & \multicolumn{3}{c}{Computational Complexity (FLOPS)}  \\ \cline{3-5} 
     &  & Attention Block & Transformer Encoder & Total in Encoder \\ \hline\hline
    1 & M4C per-layer & - & $(L+N+M)^2 = 28,900$ & $28,900$ \\
    2 & M4C 4-layer & - & $(L+N+M)^2\cdot4 = 115,600$ & $115,600$ \\
    3 & ours per-layer & $2\cdot N+2\cdot N+2\cdot M = 400$ & $(6+N)^2 = 3,136$ & $3,536$ \\ 
    4 & ours 8-layer & $2\cdot N+2\cdot N+2\cdot M = 400$ & $(6+N)^2\cdot8 = 3,136\cdot8 = 25,088$ & $25,488$ \\
    \bottomrule[1pt]
    \end{tabular}
    }
    }
    \caption{Computational complexity of two model encoders. $L=20$ is the length of question; $N=50$ is the number of OCR tokens; $M=100$ is the number of detected objects. Here we omit all vector dimensions $D$ for simplicity. In transformer encoder of our model, apart from the six vectors, we also input $50$ OCR tokens to compute dot product value in answer choosing. }
    \label{tab:com_para}
    \end{table*}
\end{center}

\section{Conclusion}
In this paper, we provide a simple but rather strong baseline for the text-based vision-and-language research. Instead of handling all modalities  over a joint embedding space or via complicated graph structural encoding, we use the vanilla attention mechanism to fuse pairwise modalities. We further split text features into two functionally different parts, \ie linguistic- and visual-part which flow into corresponding attention branch. We evaluate our simple baseline model on TextVQA, ST-VQA and TextCaps, all leading to the best performance on the public leaderboards. This sets the new state-of-the-art and our model could be the new backbone model for both TextVQA and TextCaps. What's more, we believe this work inspires a new thinking of the multi-modality encoder design. 

\newpage

\begin{figure*}[t!]
	\begin{center}
		\includegraphics[width=0.98\textwidth]{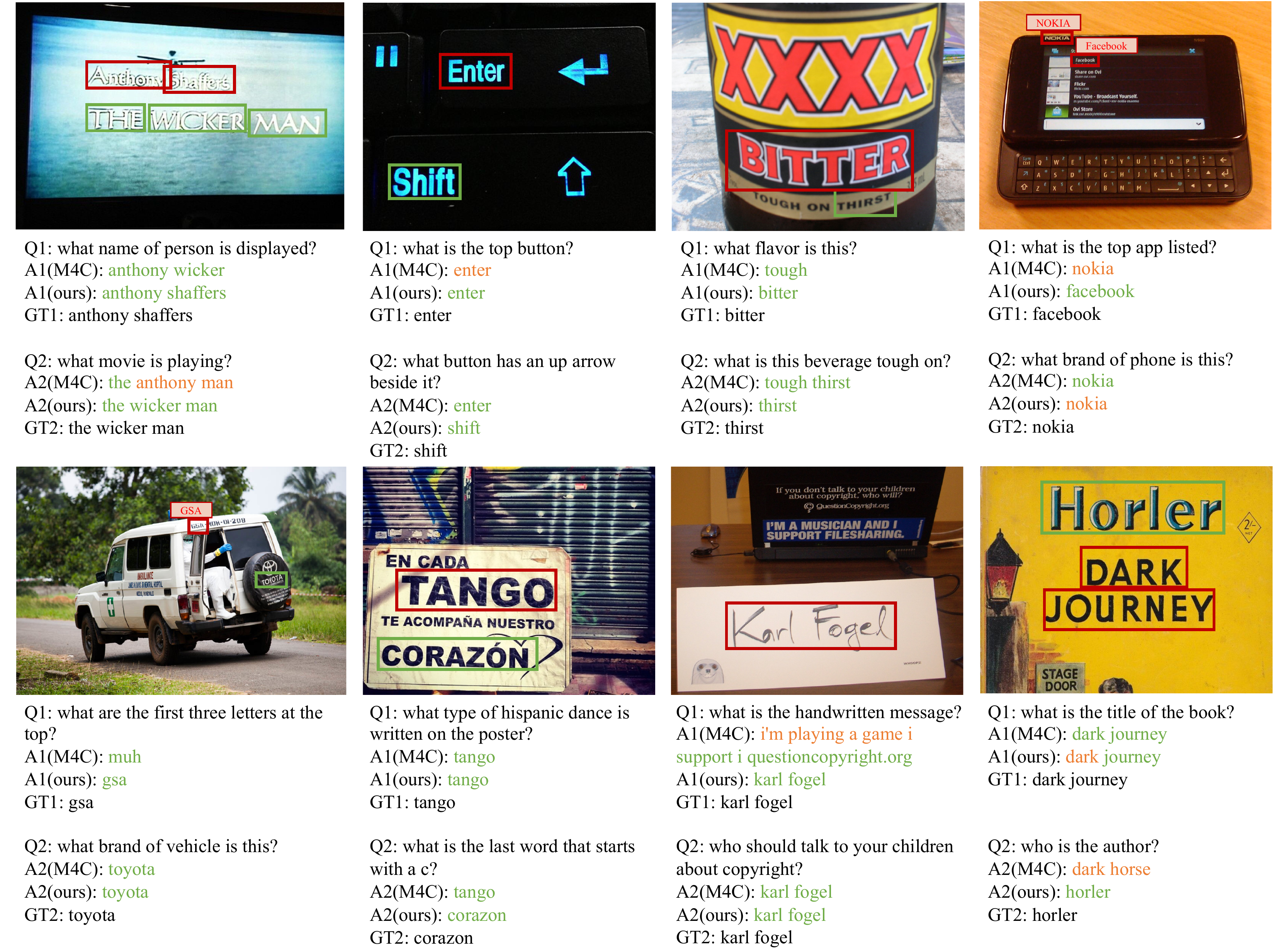}
	\end{center}
	\caption{Additional qualitative examples of our model on TextVQA. Given an image and two questions, our model can answer correctly to both. "GT" denotes Ground-Truth answer. Green words are OCR tokens and orange words are vocabulary entries. This color representation is used throughout the paper.}
	\label{fig:suc_text}
\end{figure*}

\section{Algorithm}
Algorithm~\ref{alg:textvqa} provides the pseudo-code for the process of answer generation in the TextVQA problem. The description generation process in the TextCaps problem is roughly the same, except that the decoding step $T$ is set to $30$ rather than $12$ for a longer and more expressive image description. Besides, the question features are also replaced by OCR features and Object features to guide each other.

\section{Comparison of Computation Complexity between Encoders}
As our model and M4C have the same generative decoder, we only compare encoders in Table~\ref{tab:com_para}.

In one \textit{Attention Block}, main operations are element-wise multiplication and final matrix multiplication, which yield $O(n \cdot d)=2n \cdot d$ operations, where n is number of queried features. In Transformer encoder, the complexity is $O(n^2 \cdot d)$, where $n$ is the sequence length and $d$ is the representation dimension. 

\section{Additional Experiment Details}
\noindent{\bf OCRs and Objects self-attention in TextCaps. }
Different from question self-attention in TextVQA problem that uses a BERT model to capture the sequence information, for OCRs and Objects self-attention in TextCaps we use an LSTM~\cite{hochreiter1997long} layer to capture the recurrent features of isolate regions before the self-attention calculation.

\noindent{\bf SBD-Trans training data. }
The SBD model is pretrained on a $60k$ dataset, which consists of $30,000$ images from LSVT~\cite{sun2019icdar} training set, $10,000$ images from MLT 2019~\cite{nayef2019icdar2019} training set, $5,603$ images from ArT~\cite{chng2019icdar2019} 
, and $14,859$ images selected from a bunch of datasets -- RCTW-17~\cite{shi2017icdar2017}, ICDAR 2013~\cite{karatzas2013icdar}, ICDAR 2015~\cite{karatzas2015icdar}, MSRA-TD500~\cite{yao2012detecting}, COCOText~\cite{veit2016coco}, and USTB-SV1K~\cite{yin2015multi}. The model was finally finetuned on MLT 2019 training set. 
The robust transformer based network is trained on the following datasets: IIIT 5K-Words~\cite{mishra2012scene}, Street View Text~\cite{wang2011end}, ICDAR 2013, ICDAR 2015, Street View Text Perspective~\cite{quy2013recognizing}, CUTE80~\cite{risnumawan2014robust} and ArT.

\section{Additional Qualitative Examples}
\begin{figure*}[t!]
	\begin{center}
		\includegraphics[width=0.98\textwidth]{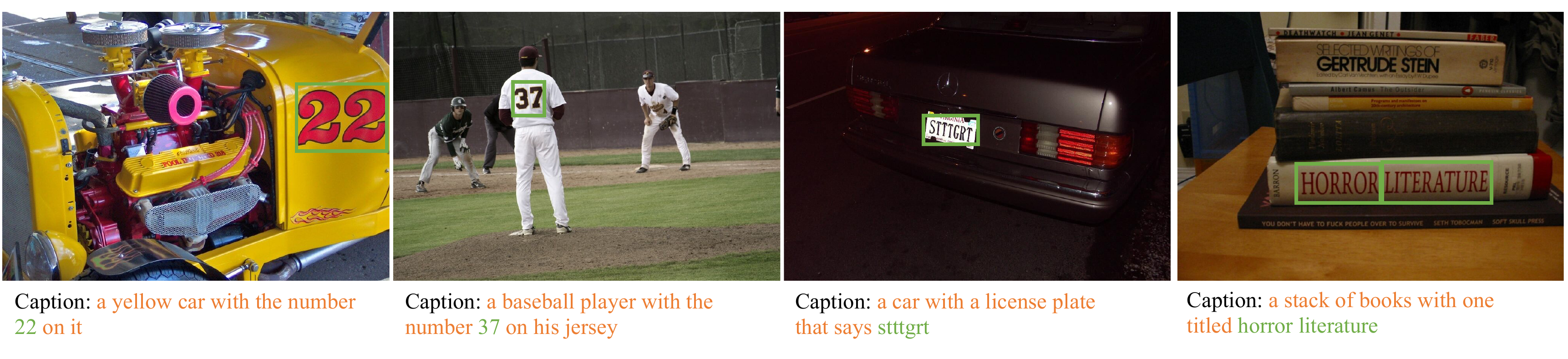}
	\end{center}
	\caption{Qualitative examples of our model on TextCaps. }
	\label{fig:suc_cap}
\end{figure*}

To demonstrate that our model not only learns fixed superficial correlation between questions and prominent OCR tokens, we present several images in the dataset with two different questions to which our model both answers correctly, while other models such as M4C fails, in Figure~\ref{fig:suc_text}. TextCaps examples are in Figure~\ref{fig:suc_cap}.


\begin{figure*}[t!]
	\begin{center}
		\includegraphics[width=0.98\textwidth]{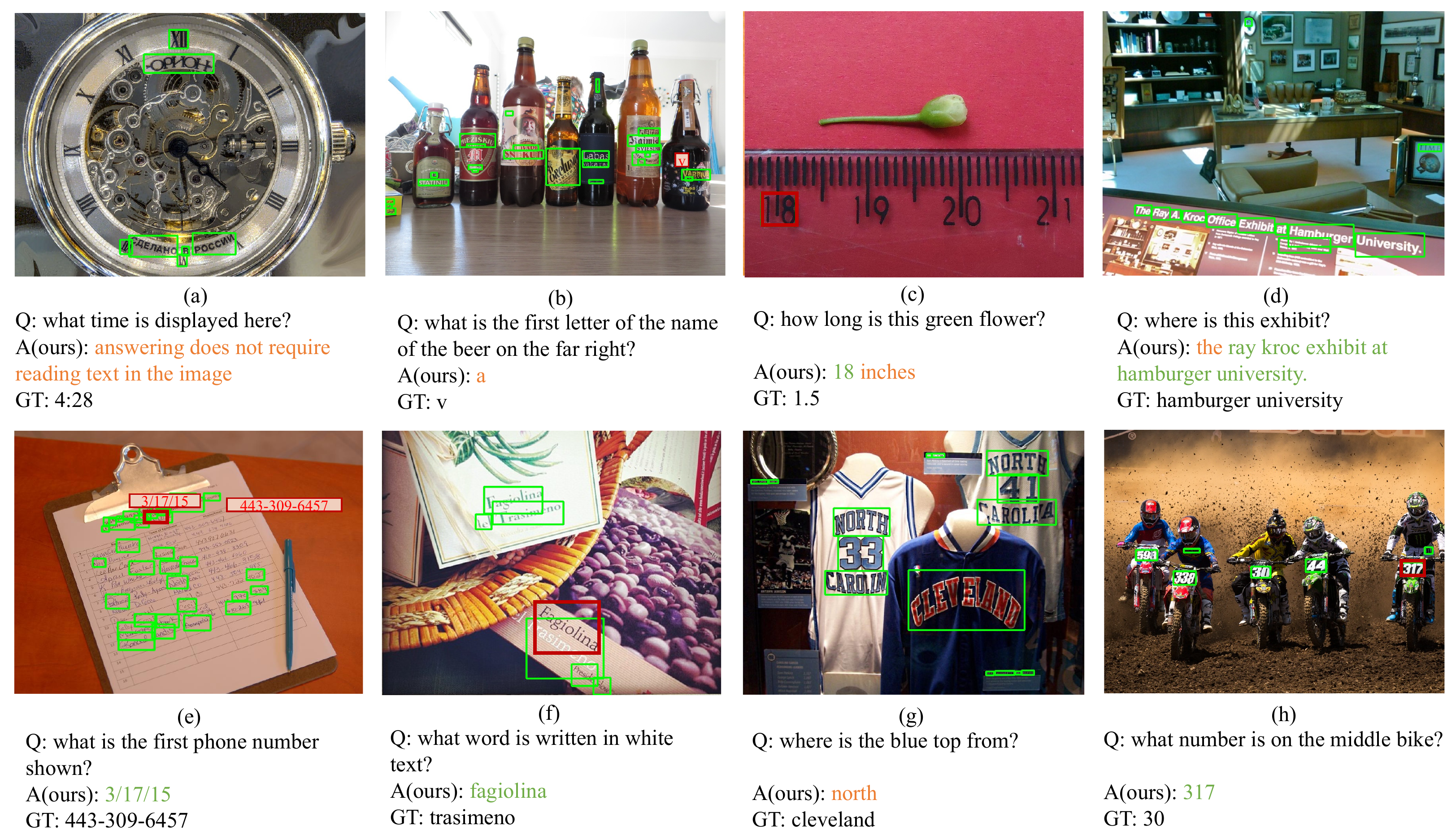}
	\end{center}
	\caption{Examples that our model fails at on TextVQA. }
	\label{fig:fa_text}
\end{figure*}

\section{Failure Cases on TextVQA}
The questions that our model struggles at can be categorized into several groups. Firstly, due to the inappropriate tricky design of the dataset, models are required to read the time, find a specific letter in a word, or read a ruler, \eg Figure~\ref{fig:fa_text}a), b) and c).
Secondly, even though our model answers reasonably, it does not lie in the region of ground-truth answers as shown in Figure~\ref{fig:fa_text}d). Thirdly, the mistake might occur because of defective reading ability that OCR tokens do not include answer candidates at all, as shown in Figure~\ref{fig:fa_text}e). Fourthly, it is difficult to discern the color of tokens as shown in Figure~\ref{fig:fa_text}f). Finally, when it comes to nuanced relationship understanding, our model is not yet complex enough to reason as shown in Figure~\ref{fig:fa_text}g) and h).

\bibliography{egbib}
\end{document}